\documentclass[letterpaper]{article} 
\usepackage{aaai2026}  
\usepackage{fancyhdr} 

\nocopyright 
\usepackage{times}  
\usepackage{helvet}  
\usepackage{courier}  
\usepackage[hyphens]{url}  
\usepackage{graphicx} 
\usepackage{natbib}  
\usepackage{caption} 
\usepackage{algorithm}
\usepackage{algorithmic}

\usepackage{amsmath}
\usepackage{amssymb}
\usepackage{mathtools}
\usepackage{booktabs}
\usepackage{multirow} 
\usepackage{tabularx}
\usepackage{newfloat}
\usepackage{listings}
\DeclareCaptionStyle{ruled}{labelfont=normalfont,labelsep=colon,strut=off} 
\floatstyle{ruled}
\newfloat{listing}{tb}{lst}{}
\floatname{listing}{Listing}
\title{RAPTOR: Real-Time High-Resolution UAV Video Prediction with \\ Efficient Video Attention}
\author{
    Zhan Chen\textsuperscript{\rm 1,2,3},
    Zile Guo\textsuperscript{\rm 1,2,3},
    Enze Zhu\textsuperscript{\rm 1,2,3},
    Peirong Zhang\textsuperscript{\rm 1,2,3},
    \\
    Xiaoxuan Liu\textsuperscript{\rm 1,2},
    Lei Wang\textsuperscript{\rm 1,2},
    Yidan Zhang\textsuperscript{\rm 1,2,3}\thanks{Corresponding author.}
}
\affiliations{
    \textsuperscript{\rm 1}Aerospace Information Research Institute, Chinese Academy of Sciences\\
    \textsuperscript{\rm 2}Key Laboratory of Target Cognition and Application Technology (TCAT)\\
    \textsuperscript{\rm 3}School of Electronic, Electrical and Communication Engineering, University of Chinese Academy of Sciences\\

    \{chenzhan21, guozile25, zhuenze23, zhangpeirong22, zhangyidan19\}@mails.ucas.ac.cn,
    \\
    \{liuxiaoxuan, wanglei002931\}@aircas.ac.cn
}

\usepackage{bibentry}

\begin{document}

\maketitle
\thispagestyle{fancy} 
\fancyhf{} 

\begin{abstract}
Video prediction is plagued by a fundamental trilemma: achieving high-resolution and perceptual quality typically comes at the cost of real-time speed, hindering its use in latency-critical applications. This challenge is most acute for autonomous UAVs in dense urban environments, where foreseeing events from high-resolution imagery is non-negotiable for safety. Existing methods, reliant on iterative generation (diffusion, autoregressive models) or quadratic-complexity attention, fail to meet these stringent demands on edge hardware. To break this long-standing trade-off, we introduce RAPTOR, a video prediction architecture that achieves real-time, high-resolution performance. RAPTOR’s single-pass design avoids the error accumulation and latency of iterative approaches. Its core innovation is Efficient Video Attention (EVA), a novel translator module that factorizes spatiotemporal modeling. Instead of processing flattened spacetime tokens with $O((ST)^2)$ or $O(ST)$ complexity, EVA alternates operations along the spatial (S) and temporal (T) axes. This factorization reduces the time complexity to $O(S + T)$ and memory complexity to $O(max(S, T))$, enabling global context modeling at $512^2$ resolution and beyond, operating directly on dense feature maps with a patch-free design. Complementing this architecture is a 3-stage training curriculum that progressively refines predictions from coarse structure to sharp, temporally coherent details. Experiments show RAPTOR is the first predictor to exceed 30 FPS on a Jetson AGX Orin for $512^2$ video, setting a new state-of-the-art on UAVid, KTH, and a custom high-resolution dataset in PSNR, SSIM, and LPIPS. Critically, RAPTOR boosts the mission success rate in a real-world UAV navigation task by 18\%, paving the way for safer and more anticipatory embodied agents.
\end{abstract}

\begin{links}
    \link{Code}{https://github.com/Thelegendzz/RAPTOR}
\end{links}

\section{Introduction}
\label{sec:intro}

\begin{figure*}[ht!]
\centering
\includegraphics[width=\linewidth]{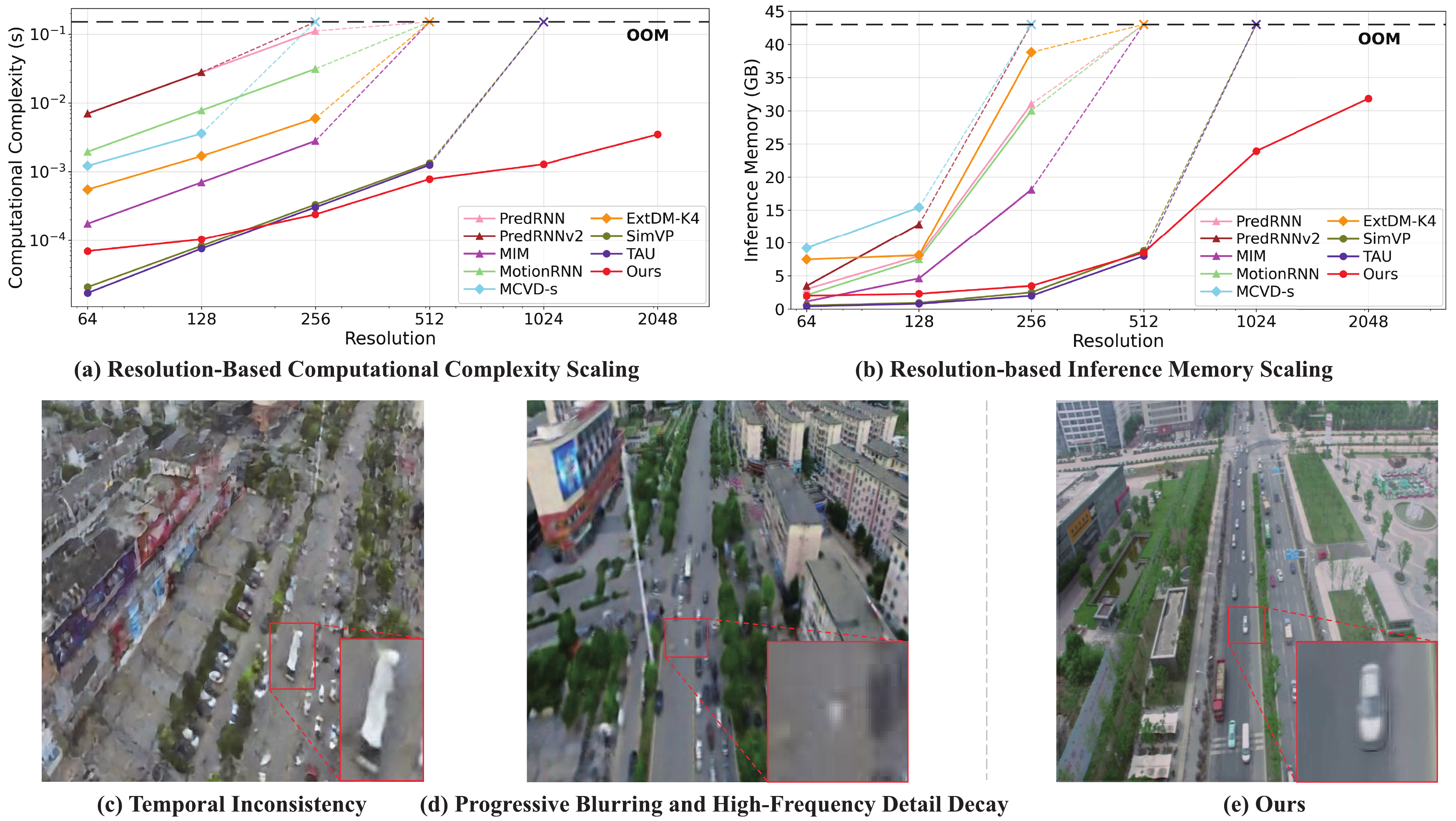}
\caption{RAPTOR enables scalable, high-fidelity video prediction for demanding, high-resolution scenarios. \textbf{(a, b)} Benchmarked on an NVIDIA RTX 6000 Ada (48GB), RAPTOR's linear complexity design maintains a low memory footprint, making it the \textbf{first framework to efficiently scale beyond $1024^2$ resolution} as competitors hit memory ceilings (Out-of-Memory). \textbf{(c, d)} In contrast, conventional methods relying on pixel-wise losses often result in \textbf{(c)} temporal tearing and \textbf{(d)} edge blurring that obscures vital targets. \textbf{(e)} The proposed RAPTOR avoids these pitfalls, generating sharp and coherent predictions that preserve crucial details for reliable downstream planning.}
\label{fig:intro}
\end{figure*}

The ability to anticipate the future is a hallmark of intelligence and a critical component for autonomous agents like UAVs navigating dynamic environments. Video prediction, the task of synthesizing future frames from past observations, offers this foresight. Yet, its practical application has been hamstrung by a fundamental trilemma: a persistent trade-off among inference speed, input resolution, and prediction quality. For a UAV, this is not an abstract challenge; it is a direct barrier to safety. To avoid a suddenly opening door or a pedestrian stepping into its path, the agent must process high-resolution video and generate sharp predictions in milliseconds. This exposes three critical bottlenecks in prior art:

\textbf{1. Prohibitive Latency.} State-of-the-art quality is often achieved by iterative models like diffusion or autoregressive RNNs \cite{voleti2022mcvd, wang2022predrnn}. Requiring hundreds of sequential steps, their inference times are orders of magnitude too slow for the real-time control loops of UAVs on edge hardware.

\textbf{2. Inability to Scale to High Resolution.} Dominant attention-based predictors \cite{bertasius2021space} jointly model spacetime by treating video as a flat sequence of tokens. This leads to a computational complexity of $O((ST) ^2)$, where S is the number of spatial locations and T is the number of frames. This quadratic scaling makes processing high-resolution video computationally infeasible, restricting models to low-resolution inputs (e.g., $64^2 –256^2$), thereby missing small but safety-critical details.

\textbf{3. Blurry and Incoherent Motion.} Many models rely on simple pixel-wise losses (e.g., MSE, L1) \cite{gao2022simvp, tan2023tau}. These objectives average over possible futures, resulting in blurry predictions and a lack of temporal consistency. This is especially problematic at high resolutions, where motion artifacts can render predictions unreliable for navigation.

To break this trilemma, we propose RAPTOR. As laid out in our title, RAPTOR is an end-to-end, single-pass architecture designed for real-time, high-resolution prediction that directly generates all future frames at once, eliminating iterative latency and error accumulation.

The cornerstone of RAPTOR is our Efficient Video Attention (EVA) module. Instead of operating on a flattened ST-length sequence with its prohibitive $O((ST) ^2)$ complexity, EVA factorizes the problem. It alternates between processing information along the temporal axis (a sequence of length T) and the spatial axis (a sequence of length S). This fundamental design choice reduces the time complexity to $O(S+T)$ and the memory complexity to $O(max(S,T))$, enabling efficient global modeling on high-resolution feature maps for the first time. To ensure high-fidelity outputs, we complement our architecture with a three-stage curriculum learning strategy that shifts the focus from pixel reconstruction to edge sharpness and perceptual realism.

Our contributions are as follows:
\begin{itemize}

\item \textbf{RAPTOR:} The first video prediction framework to achieve both real-time performance ($>$30 FPS on Jetson AGX Orin) and high-resolution ($512^2$ and beyond) processing. It outperforms prior art on multiple benchmarks and improves UAV navigation success rate by 18\%.

\item \textbf{Efficient Video Attention (EVA):} A novel spatiotemporal modeling module that avoids the quadratic bottleneck of standard attention by using axis factorization. This reduces complexity to $O(S+T)$, enabling scalable, global receptive fields for high-resolution video without resorting to patch tokenization.

\item \textbf{Three-Stage Curriculum Learning:} An effective training strategy that systematically improves prediction quality, yielding sharp edges and temporally coherent motion critical for downstream planning tasks.
\end{itemize}

\section{Related Work}
\label{sec:related}

\subsection{Video Prediction Paradigms}
Video prediction methods fall into three main paradigms:

\subsubsection{Autoregressive models.} predict frames sequentially with recurrent units. Early ConvLSTM networks~\cite{shi2015convolutional} and PredRNN~\cite{wang2017predrnn}, PredRNN++~\cite{wang2018predrnnpp} introduced spatiotemporal LSTMs and gradient highways. Subsequent works like MIM~\cite{wang2019mim} and MotionRNN~\cite{wu2021motionrnn} add memory or motion modules. These excel at $64^2$–$256^2$ but incur latency and memory increasing linearly with the horizon.

\subsubsection{Diffusion-based models.} adapt denoising diffusion to video. DDPMs~\cite{ho2020denoising} and DDIMs~\cite{song2021denoising} achieve high fidelity but require hundreds of iterative steps. Conditional video diffusion models like MCVD~\cite{voleti2022mcvd} and ExtDM~\cite{zhang2024extdm} improve efficiency via latent-space denoising or feature extrapolation, but remain unsuitable for real-time or ultra–high-resolution prediction.

\subsubsection{End-to-end models.} generate the entire sequence in one pass. SimVP~\cite{gao2022simvp} uses only an MSE loss to match SOTA at moderate resolutions. Variants such as TAU~\cite{tan2023tau} and DMVFN~\cite{hu2023dynamic} incorporate flow consistency or temporal modules, yet still rely on monolithic pixel losses and struggle at $1024^2$.

\subsection{Training Objectives and Curricula}
Most predictors use a \emph{single-stage} reconstruction loss (MSE or L1), often causing blurriness and temporal artifacts. PredRNN++ applies a simple curriculum for stability~\cite{wang2018predrnnpp}, but does not separate coarse reconstruction from edge sharpening or perceptual refinement. In contrast, our three-stage curriculum explicitly stages objectives from pixel reconstruction to gradient-based edge losses to perceptual (VGG) fidelity, yielding sharper and more temporally coherent outputs.

\subsection{Efficient Spatiotemporal Attention}
Vanilla self-attention scales as $\mathcal{O}((ST)^2)$ for $L=ST$ tokens, infeasible for high resolutions. Sparse/local methods (Swin~\cite{liu2021swin}, Longformer~\cite{beltagy2020longformer}) reduce interactions but miss global context; kernel/linear Transformers (Performer~\cite{choromanski2020rethinking}, RWKV~\cite{peng2023rwkv}, Mamba~\cite{gu2023mamba}) achieve $\mathcal{O}(L)$ but still face large $L$. Factorized spatiotemporal variants (TimeSformer~\cite{bertasius2021space}, ViViT~\cite{arnab2021vivit}) decouple axes but spatial token counts remain prohibitive at $512^2$–$1024^2$. These limitations motivate our EVA, which natively scales as $\mathcal{O}(S+T)$ while preserving full global context and fine spatial details.

\section{Method}
\label{sec:method}

\begin{figure*}[t]
    \centering
    \includegraphics[width=\linewidth]{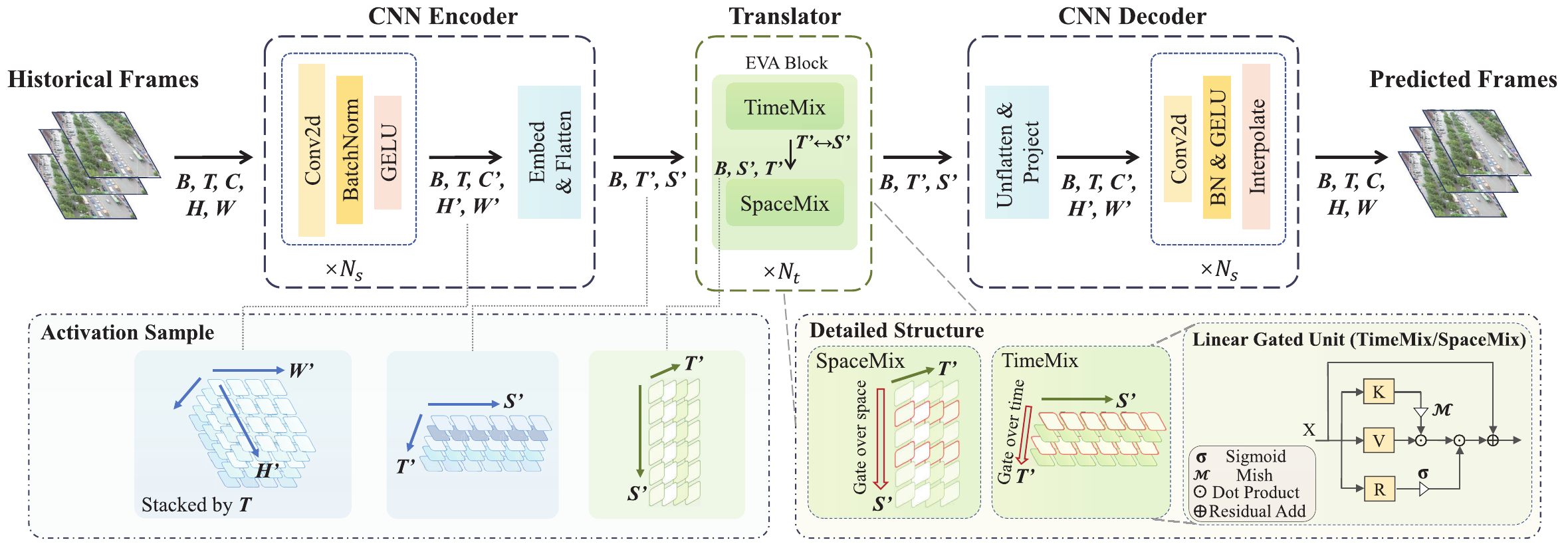}
    \caption{Overview of RAPTOR architecture. The framework follows an Encoder--Translator--Decoder design.
    The EVA Translator employs a patch-free design, operating on full feature maps from the encoder and stacking $N$ EVA blocks.
    Each block alternates a temporal \textsc{TimeMix} and a spatial \textsc{SpaceMix} built from specialized, asymmetric operators,
    achieving global spatiotemporal receptive fields with linear-in-length complexity and enabling real-time high-resolution prediction.}
    \label{fig:framework}
\end{figure*}

\subsection{Framework Overview}
\label{sec:overview}
As illustrated in Fig.~\ref{fig:framework}, RAPTOR adopts a classic Encoder--Translator--Decoder pipeline tailored for real-time, high-resolution video prediction.

\noindent\textbf{Encoder/Decoder.}
Following the efficient convolutional strategy of SimVP~\cite{gao2022simvp}, the encoder downsamples each frame and maps RGB inputs to a compact latent space. The decoder mirrors this architecture to reconstruct future frames in a single forward pass.

\noindent\textbf{Translator.}
Our primary contribution lies within the Translator---the Efficient Video Attention (EVA) stack. Given latents $\mathbf{Z}\in\mathbb{R}^{B\times T\times C'\times H'\times W'}$ from the encoder, we adopt a \textbf{fine-grained, patch-free} tokenization strategy. Instead of grouping features into coarse patches, which loses spatial resolution, we treat each location on the feature map as an individual token by flattening the spatial and channel dimensions ($S=C' \times H' \times W'$). A linear projection then maps these tokens into a fixed-width embedding $\mathbf{E}\in\mathbb{R}^{B\times T\times S}$. This dense representation, which preserves the original feature map's granularity, is then processed by the EVA stack.

\subsection{EVA: Factorized Spatiotemporal Translator}
\label{sec:eva}
\paragraph{Motivation.}
Vanilla attention over a flattened sequence of length $L{=}ST$ incurs $\mathcal{O}(L^2)$ complexity, which is prohibitive for high-resolution inputs. EVA circumvents this by transposing and alternating the modeling axes. Instead of processing the full $ST$ sequence, we apply a 1D sequence operator along the temporal ($T$) and spatial ($S$) dimensions separately. This reduces the dominant sequence length in any single pass to $\max\{T,S\}$, making the architecture highly scalable.

\paragraph{Unified Linear Gated Unit (LGU).}
A core principle of EVA's design is simplicity and consistency. 

Both \textsc{TimeMix} and \textsc{SpaceMix} are built upon a unified Linear Gated Unit (LGU), inspired by Mamba~\cite{gu2023mamba} and RWKV~\cite{peng2023rwkv} architectures.
For a given input sequence $\mathbf{U}\in\mathbb{R}^{B\times L\times S}$, the LGU operates as follows:

\begin{equation}
\small
\begin{aligned}
    \mathbf{k} = \mathrm{Mish}(\mathbf{U}\boldsymbol{W}_K),\quad
    \mathbf{v} &= \mathbf{U}\boldsymbol{W}_V,\quad
    \mathbf{r} = \sigma(\mathbf{U}\boldsymbol{W}_R) \\[4pt]
    \text{LGU}(\mathbf{U}) &= \big(\mathbf{r} \odot (\mathbf{k} \odot \mathbf{v})\big)\boldsymbol{W}_O
\end{aligned}
\label{eq:lgu}
\end{equation} 
where $\boldsymbol{W}_K, \boldsymbol{W}_V, \boldsymbol{W}_R, \boldsymbol{W}_O$ are learnable projections, and $\odot$ is the Hadamard product. This structure performs expressive, input-dependent gating with linear complexity in sequence length $L$.

\paragraph{EVA Block Structure.}
Each EVA block sequentially applies the unified LGU along the temporal and spatial dimensions, leveraging a pre-LayerNorm configuration and residual connections to enhance training stability.

First, the temporal mixing operation (\textsc{TimeMix}) is applied. Given input tensor \(\mathbf{E}\in\mathbb{R}^{B\times T\times S}\), we treat each batch independently as a 1D sequence of length $L_T = T$. To ensure causality, we apply a parameter-free temporal shift (implemented via zero-padding and rolling by one step). This shifted sequence is then processed by the LGU along the temporal axis.

Following this, the spatial mixing operation (\textsc{SpaceMix}) is performed. Starting from the output $\mathbf{E}' \in \mathbb{R}^{B \times T \times S}$ of \textsc{TimeMix}, we transpose the tensor to $\mathbf{E}'{}^\top \in \mathbb{R}^{B \times S \times T}$, interpreting each batch again as a 1D sequence—this time of length $L_S = S$. The LGU is reused here to mix information across spatial locations. This spatial mixing is fully parallel and position-wise, 
enabling efficient interaction across spatial locations without introducing directional bias or path dependencies.

By stacking $N$ such EVA blocks, the model progressively builds a global spatiotemporal receptive field, while maintaining per-block computational complexity that scales linearly with $\max\{T, S\}$.

\subsection{Complexity and Edge-Cost Model}
\label{sec:cost-model}
We evaluate per-layer inference cost in a UAV setting with $T=10$ frames at $1024\times1024$ resolution. The encoder applies $8\times$ downsampling over three stages, yielding $S=16384$ tokens per frame and $L=ST=163840$ total tokens. The latent width is $D=512$. On Jetson AGX Orin (10.65 TFLOPS FP16, 204.8 GB/s memory bandwidth), we estimate a lower bound on latency as $\text{lat}_{\text{theo}}=\max\{\text{FLOPs}/(10.65\text{ TFLOPS}),\,\text{Bytes}/(204.8\text{ GB/s})\}$.

We compare three architectures per translator layer. ViT incurs $2L^2D + 4LD^2$ FLOPs and $2L^2$ bytes for attention logits. Linear attention (RWKV/Mamba) requires $8LD^2$ FLOPs and $\mathcal{O}(LD)$ memory. EVA reduces cost to $4(T+S)D^2$ FLOPs with $\mathcal{O}(D\max\{T,S\})$ memory, by decoupling temporal and spatial modeling. 

As shown in Table~\ref{tab:orin-cost}, 
EVA achieves $1.72\times 10^{10}$ FLOPs and $\sim$1.0 GB memory per layer, with a theoretical latency of 5.1 ms—over $1600\times$ faster than ViT and $20\times$ faster than linear baselines. While others become compute-bound, EVA is memory-bound and remains in the real-time regime, making it well-suited for UAV deployment.

\begin{table}[h!]
\centering
\setlength{\tabcolsep}{2.5pt}
\begin{tabular}{lccc}
\toprule
\textbf{Method} & \textbf{FLOPs} & \textbf{Bytes} & \textbf{Latency (ms)} \\
\midrule
ViT (quadratic)      & $2.75{\times}10^{13}$ & $\sim 54$ GB   & $\sim 2600$ \\
Linear (RWKV)        & $3.44{\times}10^{11}$ & $\sim 2.4$ GB  & $\sim 32$ \\
Linear (Mamba)       & $3.92{\times}10^{11}$ & $\sim 1.8$ GB  & $\sim 37$ \\
\textbf{EVA (ours)}  & $\mathbf{1.72{\times}10^{10}}$ & $\mathbf{\sim 1.0}$ \textbf{GB} & $\mathbf{\sim 5.1}$ \\
\bottomrule
\end{tabular}
\caption{Per-layer inference cost under an $8\times$ downsampling setting ($T{=}10$, $S{=}16384$, $D{=}512$) on Jetson AGX Orin. Latency is a theoretical lower bound. ``Bytes/layer'' refers to the estimated total memory footprint.}
\label{tab:orin-cost}
\end{table}

\subsection{Three-Stage Curriculum Learning}
\label{sec:curriculum}
Pixel-wise objectives alone tend to blur edges and ignore temporal coherence, whereas optimizing all advanced objectives from scratch often destabilizes training. We therefore adopt a \textbf{coarse-to-fine} curriculum.

\paragraph{Stage I: Foundational Reconstruction.}
We first fit coarse structures using the $\ell_1$ loss:
\begin{equation}
\small
\mathcal{L}_{\text{S1}}=\|\hat{\mathbf{X}}-\mathbf{X}\|_{1}.
\end{equation}

\paragraph{Stage II: Edge and Temporal Consistency.}
We add a Gradient Difference Loss (GDL) and a frame-to-frame temporal smoothness term:

\begin{equation}
\small
\begin{aligned}
    \mathcal{L}_{\text{gdl}} =  \| \nabla_x(\hat{\mathbf{X}}) -\nabla_x(\mathbf{X}) \|_1  + \| \nabla_y(\hat{\mathbf{X}}) - \nabla_y(\mathbf{X}) \|_1 
\end{aligned}
\end{equation}

\begin{equation}
\small
\begin{aligned}
    \mathcal{L}_{\text{smooth}} =  \frac{1}{T'-1} \sum_{t=1}^{T'-1} \| \hat{\mathbf{x}}_{t+1} - \hat{\mathbf{x}}_t \|_1.
\end{aligned}
\end{equation}

\begin{equation}
\small
\begin{aligned}
    \mathcal{L}_{\text{S2}} = \mathcal{L}_{\text{S1}} &+ \lambda_{\text{gdl}}\mathcal{L}_{\text{gdl}}+ \lambda_{\text{smooth}} \mathcal{L}_{\text{smooth}}.
\end{aligned}
\end{equation}

Here $\nabla_x,\nabla_y$ are first-order difference operators.

\paragraph{Stage III: Perceptual Refinement.}
Finally we add a perceptual loss in a pretrained VGG feature space to recover high-frequency details:
\begin{equation}
\small
\mathcal{L}_{\text{S3}}=\mathcal{L}_{\text{S2}}
+\lambda_{\text{perc}}\frac{1}{T'}\sum_{t=1}^{T'}\frac{1}{S}\left\|\phi(\hat{\mathbf{x}}_{t})-\phi(\mathbf{x}_{t})\right\|_2^2,
\end{equation}
with $\phi(\cdot)$ the VGG feature extractor. We inherit weights across stages to stabilize optimization.

\paragraph{Implementation Notes.}
Our architecture incorporates several practical design choices for stability and efficiency. We use Pre-LayerNorm and residual connections in all EVA blocks. Causality in the \textsc{TimeMix} module is strictly enforced via a parameter-free \texttt{time-shift} operation on the input tensor before projection. The activation function in our \textsc{SpaceMix} LGU is Mish, which we found to perform slightly better than GELU or ReLU in preliminary tests. To manage memory for high-resolution training, we employ mixed-precision (AMP) and gradient checkpointing. These engineering details are crucial for achieving the real-time performance on edge hardware reported in our experiments.

\section{Experiments}
\label{sec:exp}

\begin{table*}[t]
\centering
\setlength{\tabcolsep}{5pt}
\begin{tabular}{l|ccc|cccc|cccc}
\toprule
\multirow{2}{*}{\bf Method} 
& \multicolumn{3}{c|}{\bf UAVid-128$^2$} 
& \multicolumn{4}{c|}{\bf UAVid-512$^2$} 
& \multicolumn{4}{c}{\bf UAVid-1024$^2$}\\
& PSNR$\uparrow$ & SSIM$\uparrow$ & LPIPS$\downarrow$ 
& PSNR$\uparrow$ & SSIM$\uparrow$ & LPIPS$\downarrow$ & FPS$\uparrow$ 
& PSNR$\uparrow$ & SSIM$\uparrow$ & LPIPS$\downarrow$ & FPS$\uparrow$ \\
\midrule
PredRNN    & 24.1 & 0.714 & 0.189 & \multicolumn{4}{c|}{\textit{OOM}} & \multicolumn{4}{c}{\textit{OOM}} \\
MIM          & 24.6 & 0.746 & 0.174 & \multicolumn{4}{c|}{\textit{OOM}} & \multicolumn{4}{c}{\textit{OOM}} \\
MotionRNN   & 26.6 & 0.757 & 0.177 & \multicolumn{4}{c|}{\textit{OOM}} & \multicolumn{4}{c}{\textit{OOM}} \\
PredRNN-V2& 25.9 & 0.761 & 0.141 & \multicolumn{4}{c|}{\textit{OOM}} & \multicolumn{4}{c}{\textit{OOM}} \\
MCVD    & 28.7 & 0.788 & 0.115 & \multicolumn{4}{c|}{\textit{OOM}} & \multicolumn{4}{c}{\textit{OOM}} \\
ExtDM      & 29.8 & 0.797 & 0.096 & \multicolumn{4}{c|}{\textit{OOM}} & \multicolumn{4}{c}{\textit{OOM}} \\
SimVP   & 28.9 & 0.791 & 0.120 & 24.5 & 0.578 & 0.120 & 31.9 & \multicolumn{4}{c}{\textit{OOM}} \\
TAU  & 29.4 & 0.805 & 0.109 & 19.8 & 0.431 & 0.499 & 31.7 & \multicolumn{4}{c}{\textit{OOM}} \\
\midrule
\textbf{RAPTOR (ours)}             & \textbf{30.7} & \textbf{0.820} & \textbf{0.076} 
& \textbf{28.4} & \textbf{0.784} & \textbf{0.095} & \textbf{145.6} 
& \textbf{19.3} & \textbf{0.455} & \textbf{0.582} & \textbf{59.6} \\
\bottomrule
\end{tabular}
\caption{UAVid $10{\to}10$ at $128^2$, $512^2$, and $1024^2$. 
FPS (batch$=$1, RTX~6000 Ada) is only reported for $512^2$ and $1024^2$ resolutions. 
\textit{OOM} (Out Of Memory) indicates peak training/inference memory $>$48\,GB. Best results in \textbf{bold}.}
\label{tab:uavid}
\end{table*}

\begin{table}[t]
\centering
\setlength{\tabcolsep}{9pt}
\begin{tabular}{l|ccc}
\toprule
\bf Method & PSNR$\uparrow$ & SSIM$\uparrow$ & LPIPS$\downarrow$ \\
\midrule
PredRNN        & 27.6 & 0.839 & 0.204 \\
MIM                & 27.9 & 0.843 & 0.177 \\
MotionRNN      & 28.3 & 0.851 & 0.169 \\
PredRNN-V2   & 28.4 & 0.838 & 0.139 \\
MCVD            & 27.5 & 0.846 & 0.104 \\
ExtDM           & 27.9 & 0.799 & 0.093 \\
SimVP             & \textbf{33.7} & 0.905 & 0.078 \\
TAU                 & 30.6 & 0.893 & 0.086 \\
\midrule
\textbf{RAPTOR (ours)}                & 32.0 & \textbf{0.912} & \textbf{0.062} \\
\bottomrule
\end{tabular}
\caption{KTH $10{\to}30$ at $64^2$. Best results in \textbf{bold}.}
\label{tab:kth}
\end{table}

We evaluate RAPTOR across four aspects: (i) comparative performance against strong state-of-the-art baselines, (ii) ablations isolating the contributions of EVA and the three-stage curriculum, (iii) scalability to ultra-high resolutions where many methods fail, and (iv) downstream impact on a real-world UAV navigation task.
\subsection{Experimental Setup}

\paragraph{Datasets.}
Our evaluation is conducted on two standard benchmarks and a custom real-world dataset. 
~\textbf{KTH}~\cite{schuldt2004recognizing} is a classic low-resolution ($64^2$) dataset for human action prediction. 
~\textbf{UAVid}~\cite{lyu2020uavid} is a challenging high-resolution dataset from UAV footage, on which we perform extensive evaluation at $128^2$, $512^2$, and a frontier $1024^2$ resolution to test the limits of scalability. 
~A custom \textbf{Real-World Navigation} dataset was collected for downstream task validation.

\paragraph{Baselines.}
To ensure a thorough comparison, we select a strong and diverse set of state-of-the-art (SOTA) methods covering three dominant paradigms: 
~\textbf{Autoregressive models} (PredRNN~\cite{wang2017predrnn}, MIM~\cite{wang2019mim}, MotionRNN~\cite{wu2021motionrnn}, and PredRNN-V2~\cite{wang2022predrnnv2}), which represent classic recurrent approaches.
~\textbf{Diffusion models} (MCVD~\cite{voleti2022mcvd} and ExtDM~\cite{zhang2024extdm}), known for their high generative fidelity.
~\textbf{End-to-end models} (SimVP~\cite{gao2022simvp} and TAU~\cite{tan2023tau}), which are the most relevant competitors in terms of efficiency.

\paragraph{Metrics and Implementation Details.}
We evaluate prediction quality using PSNR, SSIM~\cite{wang2004image}, and LPIPS~\cite{zhang2018unreasonable}. Inference speed is measured in Frames Per Second (FPS). For the downstream task, we report mission Success Rate (SR, \%). Our models were trained using PyTorch on 4* NVIDIA 6000 Ada GPUs. We leverage standard acceleration techniques including DDP, AMP, and gradient checkpointing. Further implementation details, including hyperparameters and specifics of the navigation task setup, are provided in the supplementary material.
\subsection{Quantitative Comparison with State-of-the-Art}

We present a comprehensive comparison against SOTA methods in Table~\ref{tab:uavid} and Table~\ref{tab:kth}. The results reveal a clear narrative about RAPTOR's capabilities.

On standard low-resolution benchmarks like KTH, RAPTOR demonstrates highly competitive performance. While SimVP achieves a higher PSNR, our model obtains the best SSIM (0.912) and LPIPS (0.062), indicating superior structural and perceptual quality, which are often more critical for downstream tasks than raw pixel-level accuracy.

The advantage of RAPTOR becomes decisive on high-resolution UAV data. At $128^2$, RAPTOR establishes a new SOTA across all metrics, significantly outperforming all baselines. As we scale to the challenging $512^2$ setting, this gap widens dramatically. Most prior methods fail due to Out-of-Memory (OOM) errors, highlighting a fundamental architectural bottleneck. Among the methods that can run, RAPTOR surpasses the strong SimVP baseline by a large margin (+3.9 dB PSNR, +0.206 SSIM).

Crucially, RAPTOR is the only model capable of operating at $1024^2$, establishing a new frontier for real-time video prediction. While performance metrics naturally decrease at this extreme resolution, RAPTOR's ability to process such data at interactive frame rates (59.6 FPS) is a breakthrough that no other compared method achieves. This unique scalability is a direct result of our \textbf{patch-free}, linear-complexity EVA design.

\subsection{Ablation Studies}
We conducted ablation studies to validate the individual contributions of RAPTOR's core components on the UAVid-$512^2$ dataset.

\begin{figure}[t!]
    \centering
    \includegraphics[width=\linewidth]{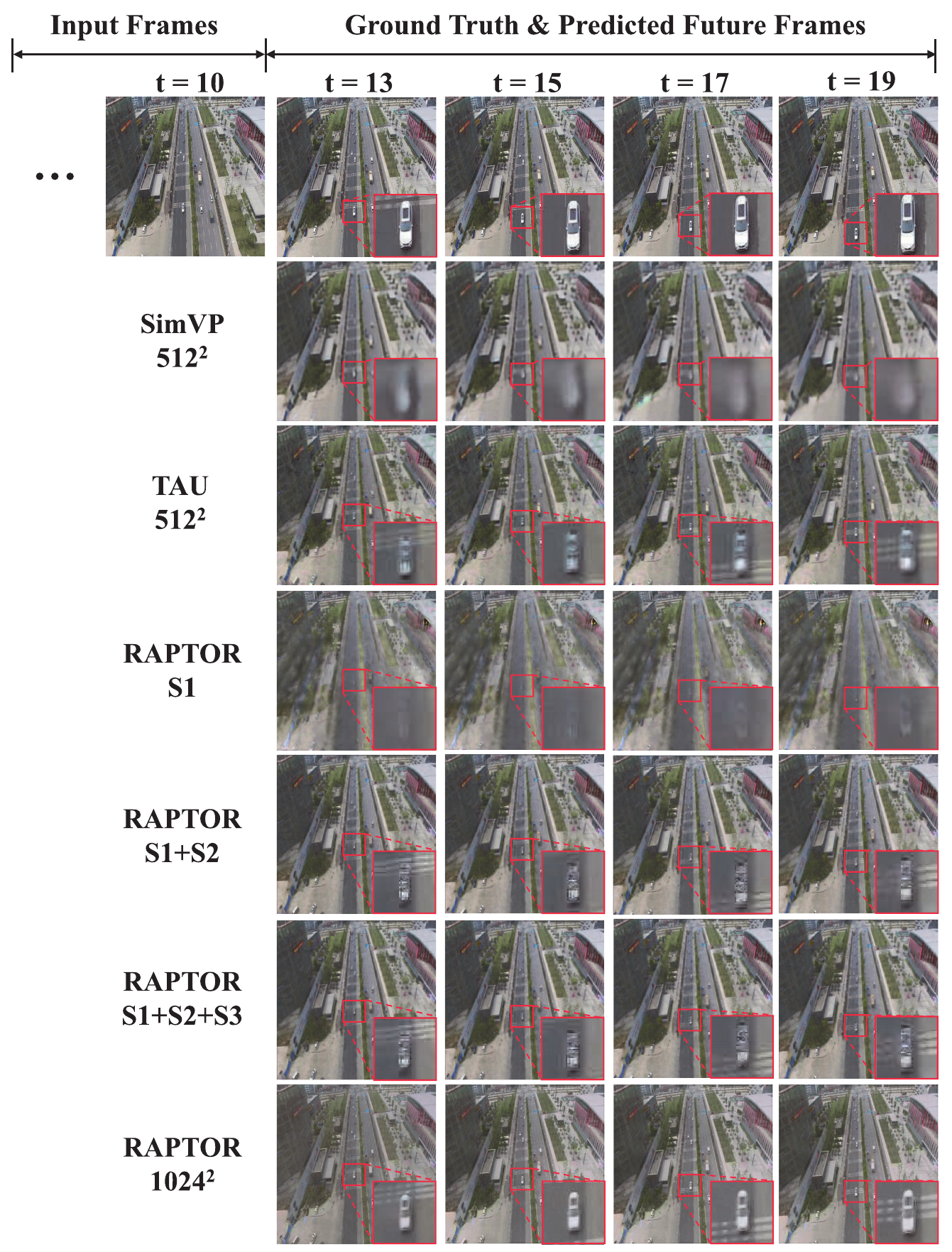}
    \caption{Qualitative comparison on a challenging UAVid scene. The magnified insets highlight how RAPTOR, particularly with the full S1+S2+S3 curriculum, generates sharper and more coherent predictions of moving vehicles compared to baselines. RAPTOR is also the only model capable of producing a prediction at $1024^2$ resolution.}
    \label{fig:qualitative_comparison}
\end{figure}

\begin{table}[h!]
\centering
\begin{tabular}{lccc}
\toprule
\textbf{Translator Variant} & PSNR$\uparrow$ & SSIM$\uparrow$ & LPIPS$\downarrow$ \\
\midrule
ViT-based & \multicolumn{3}{c}{\textit{Out Of Memory}} \\
Only TimeMix & 21.2 & 0.555 & 0.198 \\
Only SpaceMix & 27.1 & 0.713 & 0.115 \\
\textbf{RAPTOR (full EVA)} & \textbf{28.4} & \textbf{0.784} & \textbf{0.095} \\
\bottomrule
\end{tabular}
\caption{Ablation of the EVA components on UAVid-512$^2$.}
\label{tab:ablation_eva}
\end{table}

\begin{table}[h!]
\centering
\begin{tabular}{lccc}
\toprule
\textbf{Training Stage} & PSNR$\uparrow$ & SSIM$\uparrow$ & LPIPS$\downarrow$ \\
\midrule
RAPTOR (S1) & 25.31 & 0.603 & 0.121 \\
RAPTOR (S1+S2) & \bf 28.65 & \bf 0.798 & 0.102 \\
RAPTOR (S1+S2+S3) & 28.40 & 0.784 & \bf 0.095 \\
\bottomrule
\end{tabular}
\caption{Ablation of the Three-Stage Curriculum on UAVid-512$^2$, highlighting the progressive benefits of each stage.}
\label{tab:ablation_curriculum}
\end{table}

\paragraph{Efficacy of the EVA Translator.}
To prove the necessity of our factorized design, we evaluate three variants of our translator: the full EVA module, and versions using only the \textsc{TimeMix} or \textsc{SpaceMix} components. As shown in Table~\ref{tab:ablation_eva}, using only \textsc{TimeMix} fails to capture spatial details, resulting in poor performance. Using only \textsc{SpaceMix} performs better but still lags significantly behind the full model. This confirms that explicitly modeling both dimensions and, crucially, alternating between them, is essential for building a powerful spatiotemporal representation. We also note that a standard ViT-based translator runs Out-of-Memory, highlighting the efficiency of our approach.

\paragraph{Value of the Three-Stage Curriculum.}
The curriculum's effectiveness is demonstrated in Table~\ref{tab:ablation_curriculum} and visualized in Figure~\ref{fig:qualitative_comparison}. The S1 model, trained only on pixel loss, produces blurry results with low SSIM/LPIPS scores. Introducing edge and temporal losses in S2 dramatically improves structural metrics (PSNR and SSIM). The final S3, with the perceptual loss, achieves the best visual quality, as evidenced by the lowest LPIPS score. Interestingly, S3 shows a marginal drop in PSNR/SSIM compared to S2. This reflects a well-known trade-off: optimizing for perceptual realism (LPIPS) can sometimes slightly reduce pixel-level accuracy. The visual results in Figure~\ref{fig:qualitative_comparison} confirm that the S3 model is perceptually superior, justifying the full curriculum for generating visually plausible predictions.

\subsection{High-Resolution Prediction and Evaluation}
The true measure of our work lies in its ability to tackle previously intractable problems and deliver real-world value. We test RAPTOR on this frontier, first through qualitative analysis to form a hypothesis, and then through a real-world downstream task to validate it.

\paragraph{Qualitative Analysis.}
Quantitative metrics, while essential, do not capture the full picture of predictive quality. As shown in Figure~\ref{fig:qualitative_comparison}, for instance, while SimVP often scores higher than TAU on PSNR, TAU's predictions can appear more temporally consistent, highlighting the limitations of purely pixel-based evaluation.

Our qualitative results for RAPTOR tell a compelling story that aligns with our curriculum ablation study. The S1 model produces blurry outputs. The S1+S2 model, benefiting from edge and temporal losses, already generates remarkably sharp structures, especially visible in the distinct outlines of the moving vehicles. The final S1+S2+S3 model, refined by the perceptual loss, further enhances realism and textural details, resulting in the most visually plausible predictions. This progression validates our curriculum's design: S2 significantly improves structural integrity (boosting SSIM), while S3 enhances perceptual fidelity (boosting LPIPS), as confirmed by the data in Table~\ref{tab:ablation_curriculum}.

Most critically, the last row of Figure~\ref{fig:qualitative_comparison} showcases RAPTOR's breakthrough scalability. While all other methods fail at $1024^2$, RAPTOR maintains the scene's key structural and dynamic information. This leads to an insight: although our $1024^2$ predictions have lower raw metric scores than their $512^2$ counterparts (Table~\ref{tab:uavid}), they seem to preserve the fine-grained details of small, dynamic objects that are vital for navigation. This suggests a critical hypothesis: for an embodied agent, the practical value of high-resolution prediction may far outweigh its raw pixel-metric scores.

\begin{table}[h]
\centering

\setlength{\tabcolsep}{1.5pt}
\begin{tabular}{lccccc}
\toprule
\bf Method & PSNR$\uparrow$ & SSIM$\uparrow$ & LPIPS$\downarrow$ & FPS$\uparrow$ & SR$\uparrow$ \\
\midrule
GPT-4o w/o predict & - & - & - & - & 33\%\\
SimVP $512^2$ & 24.1 & 0.54 & 0.22  & 5.6 & +1\%\\
TAU $512^2$    & 18.2 & 0.42 & 0.55  & 4.9 & +4\%\\
\textbf{RAPTOR $512^2$} & \bf 27.5 & \bf 0.74 & \bf 0.11  & \bf 30.2 & \bf +7\% \\
\textbf{RAPTOR $1024^2$} & \bf 16.3 & \bf 0.49 & \bf 0.65 & \bf 18.6 & \bf +18\% \\
\bottomrule
\end{tabular}
\caption{Downstream navigation performance comparison between different types of video processing methods. SR denotes success rate (absolute improvement over the 33\% baseline). The \textit{GPT-4o w/o predict} baseline represents a reactive planner where the Vision-Language Model, GPT-4o, makes decisions based only on the current frame, without any future prediction.}
\label{tab:navigation_results}
\end{table}

\paragraph{Real-World Navigation Task.}
To rigorously test this hypothesis, we integrated RAPTOR into a real UAV's control loop for a challenging object-goal navigation task. The experiment was conducted on a university campus during nighttime to pose significant challenges for visual perception and prediction. The UAV was tasked with navigating towards periodically moving targets, including pedestrians and patrol vehicles. Due to the dynamic nature of these targets, which makes path-based metrics unreliable, we use Success Rate (SR) as the primary performance indicator, evaluated over 100 trials for each method.

\begin{figure}[t!]
    \centering
    \includegraphics[width=\linewidth]{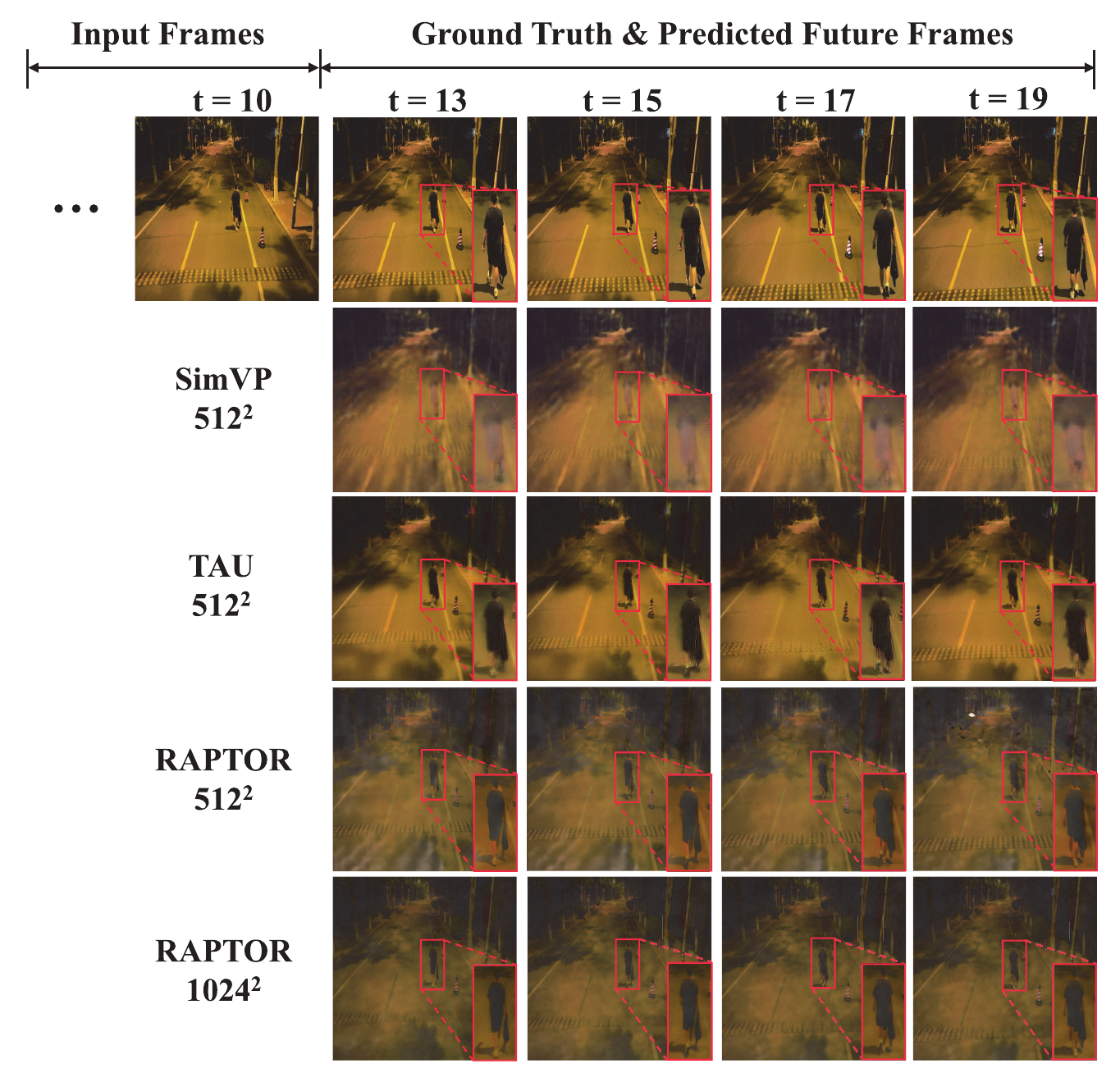}
    \caption{Qualitative comparison from the real-world UAV navigation task. Magnified insets highlight that RAPTOR (with full Three-Stage Curriculum) outperforms the baselines by generating sharp and coherent predictions of the moving pedestrian, even in a nighttime scenario.}
    \label{fig:real_world}
\end{figure}

The results, shown in Table~\ref{tab:navigation_results}, are compelling. While baselines offer minimal improvement, RAPTOR provides a substantial boost. As hypothesized, by leveraging native 1024$^2$ predictions, RAPTOR achieves a remarkable \textbf{+18\%} absolute improvement in success rate. This is because the high-resolution foresight allows the planner to perceive and react to small, distant hazards that are simply lost in downsampled inputs. This finding directly translates our model's superior predictive fidelity into tangible gains in robotic safety. Figure~\ref{fig:real_world} further provides a qualitative visualization of the real-world comparison, showing RAPTOR can predict the moving pedestrian with a clarity at both 512$^2$ and 1024$^2$ resolutions that the baselines fail to achieve.

\section{Conclusion}
\label{sec:conclusion}
In this work, we address the challenge of real-time, high-resolution video prediction for autonomous UAVs. We introduce RAPTOR, an end-to-end, single-pass framework designed to overcome the latency and scalability limitations of prior iterative methods. RAPTOR's efficacy stems from two core innovations. The first is our Efficient Video Attention (EVA) module, which tackles the quadratic complexity and memory bottleneck of processing long spatiotemporal sequences where the effective length is $L=ST$. Through spatiotemporal factorization and feature transposition, EVA applies linear gated units alternately along the spatial ($S$) and temporal ($T$) axes, which slashes the computational complexity from $O((ST)^2)$ to $O(S+T)$ and reduces the peak memory complexity to $O(\max(S,T))$. 
This fundamental reduction is what makes high-resolution processing tractable. Furthermore, its patch-free nature preserves fine-grained spatial details, while the alternating process ensures robust spatiotemporal interaction to maintain high modeling quality. The second is a three-stage curriculum learning strategy that progressively enhances the prediction's perceptual fidelity. Our extensive experiments demonstrate that RAPTOR is the first model to achieve real-time performance at $1024^2$ resolution. This capability directly translates into a significant 18\% improvement in success rate on a real-world UAV navigation task. By breaking the long-standing trade-off between resolution, speed, and quality, RAPTOR represents a significant step towards creating safer, more intelligent, and truly anticipatory aerial embodied agents.

\section{Acknowledgments}
\label{sec:acknowledgments}
This work was supported by the Key Program of Chinese Academy of Sciences under Grant RCJJ-145-24-13 and KGFZD-145-25-38, and supported by the Science and Disruptive Technology Program under Grant AIRCAS2024-AIRCAS-SDTP-03.

\bibliography{aaai2026}

\end{document}